\documentclass{article}
\usepackage{spconf,amsmath,graphicx, subcaption,multirow, cite}
\title{Towards automatic Abdominal Multi-Organ Segmentation in Dual Energy CT using cascaded 3D Fully Convolutional Network}
\name{Shuqing Chen$^{1}$ \qquad Holger Roth$^{2}$ \qquad Sabrina Dorn$^{3}$ 
     \qquad Matthias May$^{4}$ \qquad Alexander Cavallaro$^{4}$}
\secondlinename{Michael M. Lell$^{5}$ \qquad Marc Kachelrie\ss$^{3}$ \qquad Hirohisa Oda$^{2}$
     \qquad Kensaku Mori$^{2}$ \qquad Andreas Maier$^{1}$}

\address{$^{1}$ Friedrich-Alexander University Erlangen-N\"urnberg, Erlangen, Germany \\
    $^{2}$ Nagoya University, Nagoya , Japan \\
    $^{3}$ German Cancer Research Center (DKFZ), Heidelberg, Germany \\
    $^{4}$ Department of Radiology, University Hospital Erlangen, Erlangen, Germany \\
    $^{5}$ University Hospital N\"urnberg, Paracelsus Medical University, N\"urnberg, Germany
}
\begin{document}
%\ninept
%
\maketitle
“This work has been submitted to the IEEE for possible publication. Copyright may be transferred without notice, after which this version may no longer be accessible.”
\begin{abstract}
Automatic multi-organ segmentation of the dual energy computed tomography (DECT) data can be beneficial for biomedical research and clinical applications. However, it is a challenging task. Recent advances in deep learning showed the feasibility to use 3-D fully convolutional networks (FCN) for voxel-wise dense predictions in single energy computed tomography (SECT). In this paper, we proposed a 3D FCN based method for automatic multi-organ segmentation in DECT. The work was based on a cascaded FCN and a general model for the major organs trained on a large set of SECT data. We preprocessed the DECT data by using linear weighting and fine-tuned the model for the DECT data. The method was evaluated using 42 torso DECT data acquired with a clinical dual-source CT system. Four abdominal organs (liver, spleen, left and right kidneys) were evaluated. Cross-validation was tested. Effect of the weight on the accuracy was researched. In all the tests, we achieved an average Dice coefficient of 93\% for the liver, 90\% for the spleen, 91\% for the right kidney and 89\% for the left kidney, respectively. The results show our method is feasible and promising.
\end{abstract}
%In addition, manifold learning technique was applied to optimize the selection of the training dataset, the validation dataset, and the test dataset. 

%
\section{Introduction}
\label{sec:intro}
%Pelc2011
The Hounsfield unit (HU) scale value depends on the inherent tissue
properties, the x-ray spectrum for scanning and the administered contrast
media \cite{Sahani2012}. In a SECT image, materials having different elemental compositions can
be represented by identical HU values \cite{McCollough2015}.
Therefore, SECT has challenges such as limited material-specific
information and beam hardening as well as tissue characterization \cite{Sahani2012}.
DECT has been investigated to solve the challenges of SECT. In DECT, two energy-specific image data sets are acquired at two different X-ray spectra, which are produced by different energies, simultaneously. The multi-organ segmentation in DECT can be beneficial for biomedical research and clinical applications, such as material decomposition \cite{Kuchenbecker2015},
organ-specific context-sensitive enhanced reconstruction and display \cite{Dorn2017,Dorn2018}, and computation of bone mineral density \cite{wesarg2012dual}. We are aiming at exploiting the prior anatomical information that is gained through the multi-organ segmentation to provide an improved context-sensitive DECT imaging \cite{Dorn2017,Dorn2018}. The novel technique offers the possibility to present evermore complex information to the radiologists simultaneously and bears the potential to improve the clinical routine in CT diagnosis.

%clinically acquired DECT images
Automatic multi-organ segmentation on DECT images is a challenging task due to the inter-subject variance of human abdomen, the complex 3-D intra-subject variance among organs, soft anatomy deformation, as well as different HU values for the same organ by different spectra. Recent researches show the power of deep learning in medical image processing \cite{Aubreville17}. To solve the DECT segmentation problem, we use the successful experience from multi-organ segmentation in volumetric SECT images using deep learning  \cite{Roth2017,RothVisceral2017}.
The proposed method is based on a cascaded 3D FCN, a two-stage, coarse-to-fine approach \cite{Roth2017}. The first stage is used to predict the region of the interest (ROI) of the target
organs, while the second stage is learned to predict the final segmentation.
No organ-specific or energy-specific prior knowledge is required in
the proposed method. The cross-validation results showed that the proposed
method is promising to solve multi-organ segmentation problem for
DECT. To the best of our knowledge, this is the first study about
multi-organ segmentation in DECT images based on 3D FCNs. 

\section{MATERIALS AND METHODS}
\subsection{Network Architecture for DECT Prediction}
As described by Krauss et al. \cite{Krauss2011}, a mixed image display is employed in clinical practice for the diagnose using DECT. The mixed image is calculated by linear weighting of the images values of the two spectra: 

\begin{equation}
	I_{\text{mix}}=\alpha \cdot I_{\text{low}}+(1-\alpha)\cdot I_{\text{high}}
	\label{eq:DECTMixedImg}
\end{equation}

where $\alpha$ is the weight of the dual energy composition, $I_{\text{mix}}$ denotes
the mixed image. $I_{\text{low}}$ and $I_{\text{high}}$ are the images at low and high kV, respectively. 

\begin{figure*}[htbp]
    \centering
	\includegraphics[width=.8\textwidth]{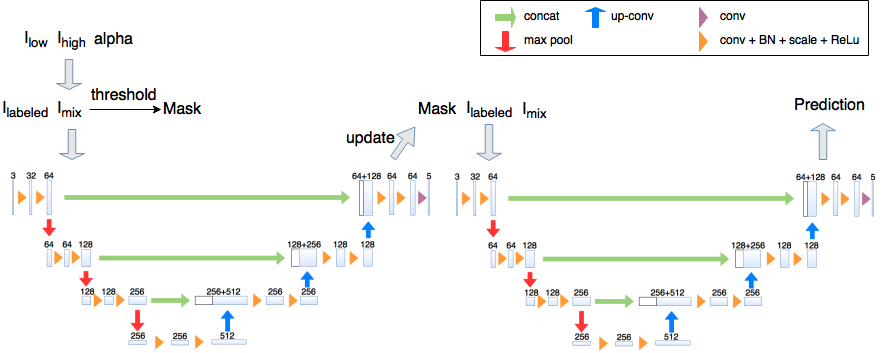}
	\vspace{-5pt}
	\caption{Cascaded network architecture for DECT multi-organ segmentation}
	\label{fig:Net}
	\vspace{-10pt}
\end{figure*}

We preprocessed the DECT images following Eq.~\ref{eq:DECTMixedImg}
straightforwardly. Figure~\ref{fig:Net} illustrates the network architecture
of the proposed method for the DECT multi-organ segmentation. First of all, mixed image is calculated by combining the images at the low energy level and the high energy level using Eq.~\ref{eq:DECTMixedImg}. Then, a binary mask is generated by thresholding the skin contour of the mixed image. Subsequently, the mixed image, the binary mask and the labeled image are given into the network as multi-channel inputs. The network consists of two stages. The first stage is applied to generate the region of the interest (ROI) in order to reduce the search space for the second stage. The prediction result of the first stage is taken as the mask for the second stage. Each stage is based on a standard 3D U-Net \cite{Cicek2016_3DUNet}, which is a fully convolutional network including an analysis and a synthesis path. We used the open-source implementation of two stages cascaded network \cite{Roth2017} developed by Roth et al. based on the 3D U-Net \cite{Cicek2016_3DUNet} and the Caffe deep learning library \cite{jia2014caffe}. A general model was trained by Roth et al. \cite{Roth2017} on a large set of SECT images including some of the major organ labels. Our model was trained by fine-tuning the general model with the mixed DECT images. The difference between the network output and the ground truth labels are compared using softmax with weight voxel-wise cross-entropy loss \cite{Cicek2016_3DUNet, Roth2017}. %formula

\subsection{Experimental Setup}
The proposed method was evaluated with 42 clinical torso DECT images scanned by the department of radiology, university hospital Erlangen. All of the images were taken from male and
female adult patients who had different clinically oriented indication justified by the radiologist. Ultravist 370 was given as contrast agent with body weight adapted volumes. The images were acquired at different X-ray tube voltage setting of 70 kV (560 mAs) and Sn 150 kV (140 mAs, with Sn filter) using a Siemens SOMATOM Force CT system with Stellar detector, an energy integrating detector. Each volume consists of 992-1290 slices of 512x512 pixels. The voxel dimensions are [0.6895-0.959, 0.6895-0.959, 0.6] mm. Four abdominal organs were tested, including liver, spleen, right and left kidneys. Ground truth was generated by experts in an inter-observer way. Training data, validation data, and test data were selected randomly with the ratio 5:1:1, i.e.\ in each test we used 6 images for validation, 6 images for test, and 30 images for training. 

%(Siemens Healthniers GmbH, Forchheim, Germany)

%To reduce the time consuming, the target organs were labeled at first using MITK application \cite{MITK} semi-automatically. Then, the coarse ground truth were refined by one expert using ITK-Snap \cite{ItkSnap} manually. To improve the accuracy of the ground truth, the refined ground truth were reviewed by different experts. 

%To avoid the bias of the data selection and to keep the dataset distribution similar,  a malnifold learning technique was applied to split the data into training dataset, validation dataset, and test dataset. First, the images were resized to the same image spacing (e.g.[3mm 3mm 5mm]). Then, the distribution of the images was calculated and plotted by using locally linear embedding (LLE) \cite{Roweis2323LLE}. Subsequently, the images were clustered into 3 classes using k-means. Finally,training data, validation data, and test data were selected randomly from these classes with the ratio 5:1:1, \ie in each test we used 2 images from each class (6 in total) for validation, 2 images from each class for test, and the remaining 30 images for training.  
%Figure \ref{fig:DataCluster} presents the result of the dimensionality reduction and the clustering of our data (data names were ignored in order to make the illustration clearly). 
%Matlab. Toolbox {[}{]} was used for the dimensionality reduction.  

\section{RESULTS}
\subsection{Performance Estimation with Cross-Validation}
\begin{figure}
  \centering
  \includegraphics[width=.7\linewidth]{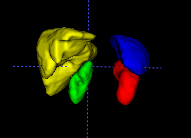}
  \vspace{-5pt}  
  \caption{3D rendering of one DECT segmentation with yellow for liver, blue for spleen, green for right kidney and red for left kidney}
  \label{fig:3DResult}
  \vspace{-10pt}
\end{figure}

\begin{figure}
  \vspace{-10pt}  
  \centering
  \includegraphics[width=\linewidth]{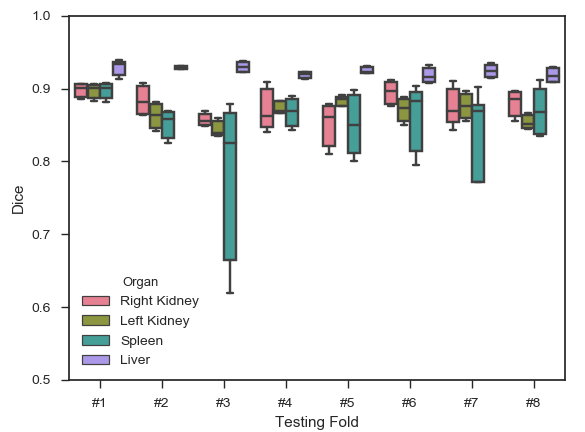}
  \vspace{-5pt}  
  \caption{Dice coefficients of the target organs with $\alpha_{\text{training}}=0.6$ and $\alpha_{\text{test}} =0.6$ for 8 different testing folds}
  \label{fig:ScenarioResult}
  \vspace{-10pt}
\end{figure}

NVIDIA GeForce GTX 1080 Ti with 11 GB memory was used for all of the experiments. The similarity between the segmentation result and the ground truth was measured with Dice metric by using the tool provided by VISCERAL \cite{TA2015}. First, the performance of the proposed method was estimated by 8-folds cross-validation, using 0.6 as $\alpha_{\text{training}}$ as well as $\alpha_{\text{test}}$. Fig.~\ref{fig:3DResult} shows one segmentation results in 3-D. Table～\ref{tab:ScenarioResult} summarizes the Dice coefficients of the segmentation results and compares DECT results with the SECT results. The proposed method under the above weight condition yielded an average Dice coefficient of 92\% for the liver, 84\% for the spleen, 88\% for the right kidney and 87\% for the left kidney, respectively. Fig. ～\ref{fig:ScenarioResult} plots the distributions of the Dice coefficients for different test scenarios and showed the high robustness of the proposed method.    
%The kidneys were segmented as accurate as SECT, the accuracy of liver and spleen of DECT was lower than SECT.
% The training took approximately 2-3 days. The execution time for the testing was much faster, cost around ... minutes.  

\begin{table}
\centering
\begin{tabular}{cccccc}%{11@{\extraclosep\fill}cccccc}
\hline
     &       & Liver & Spleen & r.Kidney & l.Kidney \\  \hline
\multirow{4}{*}{DECT}	 & Avg. &  0.92  & 0.84 &  0.88          & 0.87  \\
 & SD & 0.02  & 0.08 &  0.03          & 0.03  \\
 & Min.      & 0.84  & 0.62 &  0.80          & 0.78  \\
 & Max.      & 0.94  & 0.95 &  0.94          & 0.93  \\ \hline
\multirow{4}{*}{\begin{tabular}{@{}c@{}}SECT \\ \cite{RothVisceral2017} \end{tabular}}  & Avg. & 0.95 & 0.90 & 0.90 & 0.88 \\
 & SD & 0.02 & 0.06 & 0.06 & 0.05 \\
 & Min. 	& 0.91 & 0.75 & 0.75 & 0.78 \\
 & Max. & 0.97 & 0.95 & 0.95 & 0.94 \\ \hline
\end{tabular}
\vspace{-5pt}
\caption{Dice coefficients of cross-validation with $\alpha_{\text{training}}$=0.6 and $\alpha_{\text{test}}$ =0.6. SD is abbreviated for standard deviation. Notice that the methods used different data set, the numbers are not directly comparable.}
\label{tab:ScenarioResult}
\vspace{-10pt}
\end{table}

%\centering
%\begin{minipage}{.5\textwidth}
%  \centering
%  \includegraphics[width=.9\linewidth]{images/ClusterNolabel.png}
%  \caption{Data distribution and cluster using LLE and k-Mean}
%  \label{fig:DataCluster}
%\end{minipage}%

%\begin{figure}
%\begin{minipage}{.48\textwidth}
%  \centering
%  \includegraphics[width=.95\linewidth]{images/DICETr6Te6.eps}
%  \caption{Dice coefficients of the target organs with $\alpha_{training}=0.6$ and $\alpha_{test} =0.6$}
%  \label{fig:ScenarioResult}
%\end{minipage}
%\begin{minipage}{.48\textwidth}
%  \includegraphics[width=.95\linewidth]{images/Split1AlphaBlendingStageMax.eps}
%  \caption{Dice coefficients of target organs with alpha blending for testing fold 1}
%  \label{fig:AlphaResult}
%\end{minipage}
%\end{figure}

\subsection{Study on the Weight $\alpha$}
\begin{figure}
  \vspace{-10pt}
  \centering
  \includegraphics[width=\linewidth]{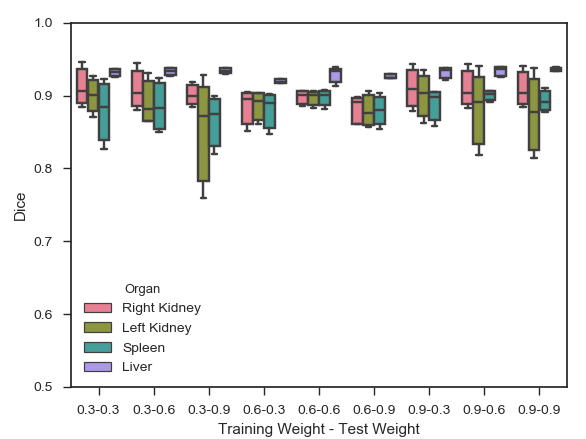}
  \vspace{-5pt}
  \caption{Dice coefficients of target organs with alpha blending for testing fold 1}
  \label{fig:AlphaResult}
  \vspace{-10pt}
\end{figure}
We are aiming at exploiting the spectral information in the DECT data. Since the $\alpha$ mixing results basically in pseudo monochromatic images comparable to single energy scans, the influence of the weight $\alpha$ on the accuracy was further researched. 0.3, 0.6 and 0.9 were chosen as $\alpha_{\text{training}}$ and $\alpha_{\text{test}}$ in this study. Fig.~\ref{fig:AlphaResult} illustrates the distributions of the Dice coefficients with different weight combination for the testing fold 1. Table~\ref{tab:AlphaResult} lists the average Dice coefficient. For all of the cases, the liver had the highest accuracy (92\%-93\%), the standard deviation of the dice coefficients (around 2\%) was fairly robust. The segmentation of the right kidney was usually more accurate than the left kidney. The best Dice values per organ per training set are highlighted in Table~\ref{tab:AlphaResult}. The test with $\alpha_{\text{training}}$=0.9 and $\alpha_{\text{test}}$=0.9 obtained the highest accuracy for liver and right kidney. The test with weight combination 0.9-0.6 showed the best segmentation for spleen, the combination with 0.9-0.3 had the finest result for left kidney. The $\alpha_{\text{test}}$=0.9 generated better segmentation for liver, the $\alpha_{\text{test}}$=0.6 worked better for spleen.

\begin{table}
\centering
\begin{tabular}{llllll}%{11@{\extraclosep\fill}cccccc}
\hline
$\alpha_{\text{training}}$-$\alpha_{\text{test}}$  & Liver & Spleen & r.Kidney & l.Kidney \\  \hline
0.3-0.3	    & 0.924  & 0.850   & \textbf{0.900}     & \textbf{0.891}  \\
0.3-0.6     & 0.925  & \textbf{0.885}   & 0.891     & 0.881  \\
0.3-0.9     & \textbf{0.926} & 0.866   & 0.872     & 0.841  \\ \hline
0.6-0.3     & 0.909  & 0.847   & 0.844     & 0.885  \\ 
0.6-0.6     & \textbf{0.922}  & \textbf{0.899}   & \textbf{0.895}     & \textbf{0.887} \\
0.6-0.9     & 0.912  & 0.872   & 0.843     & 0.873 \\  \hline
0.9-0.3     & 0.930  & 0.860   & 0.898     & \textbf{0.892} \\
0.9-0.6     & 0.932  & \textbf{0.900}   & 0.904     & 0.873 \\ 
0.9-0.9     & \textbf{0.933}  & 0.896   & \textbf{0.905}     & 0.862 \\ \hline
\end{tabular}
\vspace{-5pt}
\caption{Dice coefficients of different alpha for testing fold 1. Bold denotes the best organ results per training set.}
\label{tab:AlphaResult}
\vspace{-10pt}
\end{table}    
       
%\begin{figure*}[htbp]
%\begin{subfigure}{.35\textwidth}
%  \centering
%  \includegraphics[height=3cm,width=.95\textwidth]{images/ScenarioTr6Te3.png}
%  \caption{$\alpha_{Training}=0.6$, $\alpha_{Test}=0.3$}
%\end{subfigure}%
%\begin{subfigure}{.35\textwidth}
%  \centering
%  \includegraphics[height=3cm,width=.95\textwidth]{images/ScenarioTr6Te6.png}
%  \caption{$\alpha_{Training}=0.6$, $\alpha_{Test}=0.6$}
%\end{subfigure}%
%\begin{subfigure}{.35\textwidth}
%  \centering
%  \includegraphics[height=3cm,width=.95\textwidth]{images/ScenarioTr6Te9.png}
%  \caption{$\alpha_{Training}=0.6$, $\alpha_{Test}=0.9$}
%\end{subfigure}%
%\caption{Average Dice coefficients of target organs with alpha blending for 8 test scenarios} 
%\label{fig:}
%\end{figure*}

% To start a new column (but not a new page) and help balance the last-page
% column length use \vfill\pagebreak.
% -------------------------------------------------------------------------
%\vfill
%\pagebreak

\section{DISCUSSION AND CONCLUSION}
We proposed a deep learning based method for automatic abdominal multi-organ segmentation in DECT. The evaluation results show the feasibility of the proposed method. Compared to the results of the SECT images reported by Roth et al. \cite{RothVisceral2017}, our method is promising and robust (see Table~\ref{tab:ScenarioResult}). The segmentation of liver and spleen was less accurate than the SECT. The third testing fold had a large deviation. The reason could be that our image data were taken from patients with different disease (liver tumor, spleen tumor, etc.). The disease type is not considered by the data selection. Training and test with inconsistent symptoms could have an impact on the accuracy. 

The study on the weight can be divided into three groups with different $\alpha_{\text{training}}$. $\alpha$=0.9 is close to the low energy images which have on average the best soft-tissue contrast, $\alpha_{\text{training}}$=0.9 worked thus better in general. $\alpha$=0.6 is close to $\alpha$=0.5 which is the optimal fusion of both images with respect to signal-to-noise ratio (SNR), $\alpha_{\text{training}}$=0.6 had therefore usually the smallest deviation and showed the strongest adaptability in the inter-group comparison. The intra-group comparison showed that the cases with identical training and test conditions had a higher probability to get the best segmentation result. This is expected because the mixed images generated by the matched training and test conditions may have the highest similarity. Furthermore, the comparison of the case 0.3-0.9 (low-contrast model for high-contrast image) with the case 0.9-0.3 (high-contrast model for low-contrast image) showed that using a model trained on high-contrast images for segmenting low-contrast test images works better. In addition, liver is well segmented in middle to high $\alpha$ ranges. Spleen is segmented best at $\alpha$=0.6. Kidneys work best in matched training and test conditions. This suggests that there is an optimal $\alpha$ for each organ for image segmentation. 

The weight $\alpha$ for the mixed image calculation is currently a user-defined parameter in the preprocessing in our approach. It can be used to augment the data for the training in future. Also, the net could be modified with two image inputs. Furthermore, more organs and more scans from different patients could be used.\\
%Furthermore, our previous work \cite{ChenBVM17} showed the feasibility of multi-organ segmentation using probabilistic atlas registration. It could be possible to use the confidence map after the atlas registration as the ROI of the stage 2 instead of the stage 1.  

\textbf{ACKNOWLEDGMENTS}
This work was supported by the German Research
Foundation (DFG) through research grant No. KA 1678/20, LE 2763/2-1 and MA 4898/5-1. 

% References should be produced using the bibtex program from suitable
% BiBTeX files (here: strings, refs, manuals). The IEEEbib.bst bibliography
% style file from IEEE produces unsorted bibliography list.
% -------------------------------------------------------------------------
\bibliographystyle{IEEEbib}
\bibliography{ISBI}
%\bibliography{/Users/Shuki/Documents/Data/PhD/Publications/ChenPhD,/Users/Shuki/Documents/Data/PhD/Publications/DeepLearning}

\end{document}